# Probabilistic Inference in Influence Diagrams

**Nevin Lianwen Zhang**

Department of Computer Science, Hong Kong University of Science & Technology

lzhang@cs.ust.hk

## Abstract

This paper is about reducing influence diagram (ID) evaluation into Bayesian network (BN) inference problems. Such reduction is interesting because it enables one to readily use one's favorite BN inference algorithm to efficiently evaluate IDs. Two such reduction methods have been proposed previously (Cooper 1988, Shachter and Peot 1992). This paper proposes a new method. The BN inference problems induced by the mew method are much easier to solve than those induced by the two previous methods.

## 1 Introduction

Influence diagrams (IDs) (Howard and Matheson 1984) are a popular framework for decision analysis. An ID is an acyclic graph with three types of nodes: random nodes, decision nodes, and a single value node. Each random node is associated with a conditional probability table (CPT) and the value node is associated with a utility function. Evaluating an ID means finding an optimal decision rule for each of its decision nodes.

IDs without decision and value nodes are called Bayesian networks[1] (BNs) (Pearl 1988). They are widely used by AI researchers as a knowledge representation framework for reasoning under uncertainty. There is a rich collection of exact and approximate algorithms for inference in BNs. This paper is about how to reduce ID evaluation into BN inference problems that are as easy to solve as possible. Such reduction is interesting because it enables one to readily use one's favorite BN inference algorithm to efficiently evaluate IDs.

Cooper (1988) initiated research in this direction. He proposed a transformation of an ID into a BN and showed that optimal decision rules can be found by posing an appropriate sequence of queries to the BN. Several improvements were later introduced by Shachter and Peot (1992). This paper proposes a new method. The BN inference problems induced by the new method are much easier to solve than those induced by the two previous methods.

There are several algorithms that evaluate IDs directly without the reduction into BN inference problems (Shachter 1986, Shenoy 1992, Ndilikilikesha 1994, and Jensen *et al* 1994). We call them *direct evaluation algorithms*. An method that reduces ID evaluation into BN inference problems would be unattractive if, no matter what BN inference algorithms are used, it is less efficient than the best direct evaluation algorithm. We show that the performance of our new method, when coupled with a BN inference algorithm called VE (Zhang and Poole 1996), is always within a constant factor of the performance of the best direct evaluation method and argue that it is usually more efficient.

The fact that arbitrary BN inference algorithms can used for probabilistic calculations makes our method very attractive as compared to direct evaluation algorithms. From a system development point of view, the method enables one to easily add ID evaluation capabilities to any BN inference packages. From the efficiency point of view, speeding up inference in BNs has been and still is an active research area. There are algorithms that exploit independence of causal influence (e.g. Zhang and Poole 1996) and that exploit special structures in the conditional probability tables. The new method facilitates ready incorporation of those algorithms, as well as future advances in BN inference, in ID evaluation. We are not aware of any approximate algorithms for IDs, while there is a rich collection of approximate and simulation algorithms for BNs. The new method also opens up the possibility of approximate algorithms for ID, which might be necessary in order to solve large decision problems.

---

[1] Also known as belief networks and probabilistic influence diagrams.

We will begin with definitions related to influence diagrams and a brief review of Shachter and Peot's method (Section 2). Foundations for our new method will be laid in Section 3 and details will be worked out in Section 4. The new method will be illustrated through an example in Section 5 and compared with Shachter and Peot's method and direct evaluation algorithms in Section 6. Conclusions will be drawn in Section 7.

## 2    Influence diagrams

In the original definition of IDs (Howard and Matheson 1984), there is only one value node. We allow multiple value nodes here so that separability in the utility function can be represented. See Tatman and Shachter (1990) for discussions on separability of utility functions.

IDs are required to satisfy several constraints. First, value nodes cannot have children. Second, IDs must be *regular* in the sense that there must be a directed path that contains all the decision nodes. The last decision node on the path will be referred to as the *tail decision node*. Third, they must be *no-forgetting* in the sense that a decision node and its parents be parents to all subsequent decision nodes. The rationale behind the no-forgetting constraint is that information available now should also be available later if the decision-maker does not forget.

*Value networks* refer to IDs that do not contain decision nodes. *Bayesian networks* (BNs) (Pearl 1988) are IDs that consists of only random nodes. In the following, the terms "nodes" and "variables" will be used interchangeably.

We shall use $\Omega_x$ to denote the *frame* of variable $x$, i.e. the set of possible values of $x$. For a set $X$ of variables, $\Omega_X$ stands for the Cartesian product $\prod_{x \in X} \Omega_x$.

Let $d_1, \ldots, d_k$ be all the decision nodes in an ID $\mathcal{N}$. A *decision rule* for a decision node $d_i$ is a mapping $\delta_i : \Omega_{\pi_{d_i}} \rightarrow \Omega_{d_i}$. A *policy* is a list of decision rules $\Delta = (\delta_1, \ldots, \delta_k)$ consisting of one rule for each decision node. *To evaluate an ID* is to find an *optimal policy* that maximizes the expected utility and to compute the optimal expected utility.

ID evaluation requires a lot of probabilistic calculations. This paper is concerned with identifying a set of probabilistic inference problems such that optimal decision rules can be readily obtained from their so-lutions. The problems should be as easy to solve as possible.

Cooper (1988) initiated research in this direction. Several improvements to Cooper's method were proposed by Shachter and Peot (1992). This section briefly reviews the method by Shachter and Peot.

The method applies only in the case when there is one value node. Let $\mathcal{N}$ be an ID with one value node. Denote the value node by $v$. For simplicity, assume that the utility function $f_v(\pi_v)$ of $v$ is non-negative[2]. The node is converted into a binary random node with the following conditional probability:

$$
\begin{aligned}
P(v=1|\pi_v) &= \frac{f_v(\pi_v)}{M_v}, \\
P(v=0|\pi_v) &= 1 - P(v=1|\pi_v),
\end{aligned} \tag{1}
$$

where $M_v = max_{\pi_v} f_v(\pi_v)$. This transformation will be referred to as *Cooper's transformation.*

Each decision node $d$ is also converted into a random node with the following conditional probability:

$$
P(d=\alpha|\pi_d)=1/|\Omega_d|,
$$

for each possible value $\alpha$ of $d$, where $|\Omega_d|$ is the number of possible values of $d$. After the transformations, $\mathcal{N}$ becomes a BN. Denote the BN by $\mathcal{N}'$.

According to the regularity constraint, there exists a directed path that contains all decision nodes. Let $d_1, \ldots, d_k$ an enumeration of the decision nodes in the order they appear in the path. It is shown that an optimal decision rule $\delta_k^*$ for $d_k$ can be obtained by

$$
\delta_k^*(\pi_{d_k}) = arg\ max_{d_k} P_{\mathcal{N}'}(d_k, \pi_{d_k}|v=1).
$$

After the rule has been computed, the conditional probability of $d_k$ is changed to $P_{\delta_k^*}(d_k|\pi_{d_k})$. An optimal decision rule for $d_{k-1}$ is then computed using the same formula except with $k$ replaced by $k-1$. Optimal decision rules for $d_{k-2}, \ldots, d_1$ are computed recursively in the same fashion.

Shachter and Peot's method reduces the evaluation of $\mathcal{N}$ into the following BN inference problems:

$$
P_{\mathcal{N}'}(d_k, \pi_{d_k}|v=1), P_{\mathcal{N}'}(d_{k-1}, \pi_{d_{k-1}}|v=1), \ldots, P_{\mathcal{N}'}(d_1, \pi_{d_1}|v=
$$

We will show that an ID can be evaluated by solving BN inference problems that are much easier than those listed above.

---

[2]If the utility function takes negative values, a constant can be added to it so that it takes only non-negative values. Addition of a constant to the utility function does not change the optimal policies. Moreover, the optimal expected value of an ID equals its optimal expected value after the addition of the constant minus the constant.

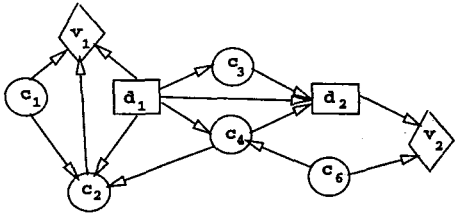

Figure 1: An ID. Random nodes are drawn as ellipses, decision nodes as rectangles, and value nodes as diamonds.

## 3    Decomposition theorem

Suppose $\mathcal{N}$ is an ID and $d$ is the tail decision node. This section shows that $\mathcal{N}$ can be decomposed into two components, called tail and body respectively, such that an optimal decision rule for $d$ can be found in the tail and optimal decision rules for all other decision nodes can be found in the body. The body is again an ID and hence the decomposition can be repeated in the body.

### 3.1    Downstream and upstream sets

We begin by partitioning the set of nodes in $\mathcal{N}$ into several subsets w.r.t the tail decision node $d$. The *moral graph* of an ID is obtained by first adding undirected edges between pairs of parents of each node so that they are pairwise connected and then dropping directions of all the directed edges. Let $x$ and $y$ be two nodes and $S$ be a set of nodes that does not contain $x$ or $y$. We say that $S$ *m-separates* $x$ and $y$ if, in the moral graph, every path connecting them contains at least one node in $S$.

Let $X$ be the set of all nodes in an ID $\mathcal{N}$. The *upstream set of $\mathcal{N}$ w.r.t to $d$*, denoted by $X_1$, is the set of nodes in $X \backslash \pi_d$ that are m-separated from $d$ by $\pi_d$. The *downstream set of $\mathcal{N}$ w.r.t $d$*, denoted by $X_2$, is the set of nodes in $X \backslash \pi_d$ that are not m-separated from $d$ by $\pi_d$. Note that $d \in X_2$ and that the three sets $X_1$, $\pi_d$ and $X_2$ constitute a partition of the set $X$.

Define $\pi_{d,2}$ be the set of nodes in $\pi_d$ that have at least one parent in $X_2$ and set $\pi_{d,1} = \pi_d \backslash \pi_{d,2}$. The four sets $X_1$, $X_2$, $\pi_{d,1}$, and $\pi_{d,2}$ constitute another partition of $X$.

Consider the ID in Figure 1. The set of parents of $d_2$ is $\pi_{d_2} = \{d_1, c_3, c_4\}$ and the downstream set $X_2$ w.r.t $d_2$ is $X_2 = \{d_2, c_6, v_2\}$. Since $c_4$ the only parent of $d_2$ that has a parent in the downstream set, $\pi_{d_2,2} = \{c_4\}$. Hence $\pi_{d_2,1} = \{d_1, c_3\}$.

A node $x$ is an *ancestor* to another node $y$ if there is a directed path from $x$ to $y$. A *ancestral set $an(A)$* of a set of nodes $A$ consists of nodes in $A$ and their ancestors. The following proposition summarizes properties of the aforementioned sets.

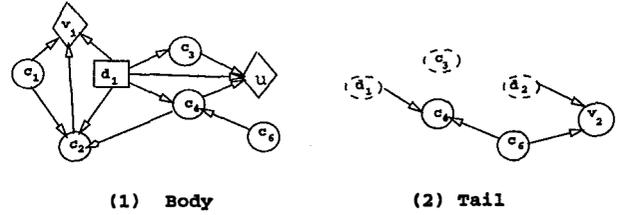

Figure 2: Tail and body: The BN in (2) is the tail of the influence diagram in Figure 2 w.r.t $d_2$; probabilities of the dashed nodes are uniform distributions. The ID in (1), with the value node $u$ ignored, is the body of the ID in Figure 2 with w.r.t $d_2$. With $u$, it is the augmented body (Subsection 3.2).

**Proposition 1** *Suppose $d$ is the tail decision node in an ID. Then (1) the node $d$ is the only decision node in the downstream set $X_2$; (2) all nodes in $\pi_{d,2}$ are random nodes; (3) all nodes in $an(\pi_{d,2}) \cap X_2$ are random nodes; and (4) all other decision nodes are in $\pi_{d,1}$.*

### 3.2    Bodies and tails

The *body of $\mathcal{N}$ w.r.t $d$* is an ID given by:

Procedure body($\mathcal{N}, d$):

1. Prune from $\mathcal{N}$ all the nodes in $X_2 \backslash an(\pi_{d,2})$.
2. Return the resulting ID.

We use $\mathcal{B}$ to denote the body. According to Proposition 1 (4), for any decision node $d' \neq d$, $d'$ is in $\mathcal{B}$ and it has the same parents in $\mathcal{B}$ as in $\mathcal{N}$.

Define the *tail of $\mathcal{N}$ w.r.t $d$* by the following procedure:

Procedure tail($\mathcal{N}, d$):

1. Prune from $\mathcal{N}$ all nodes in $X_1$.
2. Prune arcs into $d$ and nodes in $\pi_{d,1}$.
3. Prune conditional probabilities of random nodes in $\pi_{d,1}$.
4. For each node $x \in \{d\} \cup \pi_{d,1}$, set $P_{\mathcal{T}}(x) = 1/|\Omega_x|$, where $|\Omega_x|$ is the number of possible values of $x$.
5. Convert all the value nodes in $X_2$ into random nodes by Cooper's transformation.
6. Return the resulting BN.

We use $\mathcal{T}$ to denote the tail. It is a BN for the following reasons. It consists the decision node $d$, nodes in $\pi_{d,1}$, nodes in $\pi_{d,2}\cup X_2\backslash\{d\}$. The node $d$ and nodes in $\pi_{d,1}$ are associated with uniform distributions. According to Proposition 1, nodes in $\pi_{d,2}\cup X_2\backslash\{d\}$ are either random nodes or value nodes. Random nodes in the set inherit their conditional probabilities from $\mathcal{N}$, while conditional probabilities for value nodes in the set are obtained from their utility functions via Cooper's transformation. So all nodes in $\mathcal{T}$ are associated with probabilities and hence $\mathcal{T}$ is a BN.

Let $V_2$ be the set of all value nodes in the tail. Define the *evaluation functional* $e_\mathcal{T}(\pi_d, d)$ of the tail $\mathcal{T}$ by

$$e_\mathcal{T}(\pi_d, d) = \sum_{v\in V_2} P_\mathcal{T}(v{=}1|\pi_d, d)M_v. \tag{2}$$

### 3.3  Decomposition theorem

Define the *augmented body of* $\mathcal{N}$ *w.r.t* $d$ by the following procedure:

Procedure $\mathrm{augBody}(\mathcal{N}, d, e_\mathcal{T}(\pi_d, d))$:

1. $\mathcal{B}{=}\mathrm{body}(\mathcal{N}, d)$.
2. Introduce a new value node $u$ to $\mathcal{B}$. Make it a child of each node in $\pi_d$ and set its utility function as follows:

$$f_u(\pi_u) = max_d e_\mathcal{T}(\pi_d, d).$$

3. Return the resulting ID.

We use $\mathcal{B}$ to denote the augmented body from now on.

**Theorem 1** (Decomposition Theorem)

1. *An optimal decision rule* $\delta^*$ *for the tail decision node* $d$ *is given by*

$$\delta^*(\pi_d) = arg\ \ max_d e_\mathcal{T}(\pi_d, d). \tag{3}$$

2. *A policy* $\Delta_1$ *for the augmented body* $\mathcal{B}$ *of* $\mathcal{N}$ *w.r.t* $d$ *is optimal if and only if the policy* $(\Delta_1, \delta^*)$ *is optimal for* $\mathcal{N}$.

3. *The optimal expected value of* $\mathcal{B}$ *is the same as that of* $\mathcal{N}$.

## 4  Evaluating IDs

The decomposition theorem gives us the following procedure for evaluating an ID: (1) decompose it to two components — tail and body — w.r.t the tail decision node, (2) find an optimal decision rule for the tail decision node in the tail, and (3) repeat the process in the body. This section looks at the necessary computations in more detail and identifies the BN inference problems that one needs to solve. We also introduce several optimizations that make the BN inference problems easier to solve.

### 4.1  Simplifying computations in the tail

To obtain the evaluation functional $e_\mathcal{T}(\pi_d, d)$ of the tail $\mathcal{T}$, we need to compute the marginal probability $P_\mathcal{T}(\pi_d, d)$ and the marginal probability $P_\mathcal{T}(v{=}1, \pi_d, d)$ for each value node $v\in V_2$. This subsection shows that some of the nodes that appear in the marginal probabilities can be deleted and that some of the nodes in $\mathcal{T}$ can be pruned when computing each of the marginal probabilities.

#### 4.1.1  Irrelevant parents of decision nodes

Let $\pi_{d,i}$ be the set of nodes in $\pi_{d,1}$ that, in $\mathcal{N}$, are not parents to nodes in $\pi_{d,2}\cup X_2\backslash\{d\}$. Each $x\in\pi_{d,i}$ is an isolated node in $\mathcal{T}$ for the following reasons. First, $x$ has no parents in $\mathcal{T}$ since arcs into nodes in $\pi_{d,1}$ have been removed by $\mathtt{tail}$. Second, $x$ has no children in $\mathcal{T}$. This is because $\mathcal{T}$ consists of the node $d$, nodes in $\pi_{d,1}$, and nodes in $\pi_{d,2}\cup X_1\backslash\{d\}$. The node $d$ and nodes in $\pi_{d,1}$ cannot be children of $x$ since all arcs into them have been removed by $\mathtt{tail}$ and nodes $\pi_{d,2}\cup X_1\backslash\{d\}$ are not children of $x$ by the definition of $\pi_{d,i}$. Define the *reduced tail* of $\mathcal{N}$ w.r.t $d$ by the following procedure:

Procedure $\mathtt{redTail}(\mathcal{N}, d)$:

1. $\mathcal{T}{=}\mathtt{tail}(\mathcal{N}, d)$.
2. Prune from $\mathcal{T}$ nodes in $\pi_{d,i}$.
3. Return the resulting BN.

We use $\mathcal{T}_r$ to denote the reduced tail. It consists of nodes in $\pi_{d,r}\cup X_2$, where $\pi_{d,r} = \pi_d\backslash\pi_{d,i}$. Because each member of $\pi_{d,i}$ is an isolated node in $\mathcal{T}$ and its probability is the uniform distribution, the joint probabilities of $\mathcal{T}$ and $\mathcal{T}_r$ are related by

$$P_\mathcal{T}(\pi_d, X_2) = P_{\mathcal{T}_r}(\pi_{d,r}, X_2)\prod_{x\in\pi_{d,i}} 1/|\Omega_x|,$$

where $|\Omega_x|$ is the number of possible values of $x$. Therefore

$$e_\mathcal{T}(\pi_d, d) = \sum_{v\in V_2} \frac{P_{\mathcal{T}_r}(v{=}1, \pi_{d,r}, d)}{P_{\mathcal{T}_r}(\pi_{d,r}, d)}M_v. \tag{4}$$

Hence the evaluation functional can be computed from the marginal probability $P_{\mathcal{T}_r}(\pi_{d,r}, d)$ and the marginal probability $P_{\mathcal{T}_r}(v{=}1, \pi_{d,r}, d)$ for each value node $v\in V_2$. Those marginal probabilities involve less nodes than $P_\mathcal{T}(\pi_d, d)$ and $P_\mathcal{T}(v{=}1, \pi_d, d)$.

Equation (4) also implies that the evaluation functional does not depend on nodes in $\pi_i$ and can be rewritten as $e_\mathcal{T}(\pi_{d,r}, d)$. This fact in turn has two implications. First, the optimal decision rule for $d$ given by equation (3) does not depend on nodes in $\pi_{d,i}$. For this reason, nodes in $\pi_{d,i}$ will be called *irrelevant parents* of $d$ (Tatman and Shachter 1990) and the nodes in $\pi_{d,r}$ will be called *relevant parents* of $d$.

Second, the utility function of the new value node $u$ in the augmented body $\mathcal{B}$ does not depend on the irrelevant parents of $d$. Consequently there is no need to make $u$ a child to those nodes. From now on we assume that, in $\mathcal{B}$, $u$ is a child of only relevant parents of $d$ and its utility function of $u$ is given by

$$f_u(\pi_{d,r}) = max_d e_\mathcal{T}(\pi_{d,r}, d). \tag{5}$$

We will use $\mathcal{T}$ to denote the reduced body from now on.

### 4.1.2   Pruning irrelevant nodes

For any $v \in V_2$, consider $P_\mathcal{T}(v=1, \pi_{d,r}, d)$. It is well known that nodes outside $an(\pi_{d,r} \cup \{d, v\})$ are irrelevant to the marginal probability (e.g. Shachter 1988). Let $\mathcal{T}_v$ be the BN obtained from $\mathcal{T}$ by pruning nodes outside $an(\pi_{d,r} \cup \{d, v\})$. Then $P_\mathcal{T}(v=1, \pi_{d,r}, d) = P_{\mathcal{T}_v}(v=1, \pi_{d,r}, d)$.

Similarly, let $\mathcal{T}'$ be the BN obtained from $\mathcal{T}$ by pruning nodes outside $an(\pi_{d,r} \cup \{d\})$. Then $P_\mathcal{T}(\pi_{d,r}, d) = P_{\mathcal{T}'}(\pi_{d,r}, d)$. Consider the node $d$. It has no parents in $\mathcal{T}$ and hence has no parents in $\mathcal{T}'$. Children of $d$ in $\mathcal{T}$ cannot be in the ancestral set $an(\pi_{d,r} \cup \{d\}$, for otherwise there would directed loops in the original ID. Hence children of $d$ are not in $\mathcal{T}'$. Therefore $d$ is an isolated node in $\mathcal{T}'$. Let $\mathcal{T}_c$ be obtained from $\mathcal{T}'$ by pruning the isolated node $d$. Since the probability of $d$ is the uniform distribution, $P_{\mathcal{T}'}(\pi_{d,r}, d) = P_{\mathcal{T}_c}(\pi_{d,r})/|\Omega_d|$.

For any BN $\mathcal{M}$ and any subset $A$ of nodes in $\mathcal{M}$, let $\mathtt{BNinf}(A, \mathcal{M})$ be a procedure that computes the marginal probability $P_\mathcal{M}(A)$. Arbitrary BN inference algorithms can be used in the procedure. According to the foregoing discussions, the evaluation functional $e_\mathcal{T}(\pi_{d,r}, d)$ can be obtained by using the following procedure:

Procedure $\mathtt{evalFun}(\mathcal{T}, d)$

1. Obtain $\mathcal{T}_c$ from $\mathcal{T}$ by pruning nodes outside $an(\pi_{d,r})$. Compute $P_{\mathcal{T}_c}(\pi_{d,r})$ by calling $\mathtt{BNinf}(\pi_{d,r}, \mathcal{T}_c)$.

2. For each $v \in V_2$, obtain $\mathcal{T}_v$ by pruning nodes outside $an(\pi_{d,r} \cup \{d, v\})$. Compute $P_{\mathcal{T}_v}(v=1, \pi_{d,r}, d)$ by calling $\mathtt{BNinf}((v=1, \pi_{d,r}), \mathcal{T}_v)$

3. Set

$$e_\mathcal{T}(\pi_{d,r}, d) = \frac{\sum_{v \in V_2} P_{\mathcal{T}_v}(v=1, \pi_{d,r}, d)}{P_{\mathcal{T}_c}(\pi_{d,r})/|\Omega_d|}.$$

4. Return $e_\mathcal{T}(\pi_{d,r}, d)$ and $P_{\mathcal{T}_c}(\pi_{d,r})$.

Note that the following BN inference problems are solved:

$$P_{\mathcal{T}_c}(\pi_{d,r}), P_{\mathcal{T}_v}(v=1, \pi_{d,r}, d) \text{ for each } v \in V_2.$$

Also note that in addition to the evaluation functional, $\mathtt{evalFun}$ also returns the marginal potential $P_{\mathcal{T}_c}(\pi_{d,r})$. It will be used in the next subsection.

### 4.2   Simplifying computations in the body

The augmented body $\mathcal{B}$ contains nodes in the set $an(\pi_{d,2}) \cap X_2$. The set is empty when no nodes in $\pi_d$ have parents in $X_2$, i.e. when $\pi_{d,2} = \emptyset$. This subsection is concerned with the case when the set is not empty and shows that nodes in the set can be pruned from $\mathcal{B}$. Pruning nodes from $\mathcal{B}$ simplifies computations in the body.

Pruning nodes in $an(\pi_{d,2}) \cap X_2$ from $\mathcal{B}$ requires conditional probabilities of some of the remaining nodes be changed. The changes can be made with little numerical computation by using the marginal probability $P_{\mathcal{T}_c}(\pi_{d,r})$. Since the marginal probability must be computed in order to obtain the evaluation functional of the reduced tail, the cost of node pruning is small.

The resulting ID after pruning $an(\pi_{d,2}) \cap X_2$ from $\mathcal{B}$ will be called the *reduced body* of $\mathcal{N}$ w.r.t $d$. Formally, it is obtained from the augmented body $\mathcal{B}$, the reduced tail $\mathcal{T}$, and the marginal probability $P_{\mathcal{T}_c}(\pi_{d,r})$ via the following procedure:

Procedure $\mathtt{redBody}(\mathcal{B}, P_{\mathcal{T}_c}(\pi_{d,r}), \mathcal{T})$:

1. Prune from $\mathcal{B}$ all nodes in $an(\pi_{d,2}) \cap X_2$.

2. Prune arcs into and conditional probabilities of nodes in $\pi_{d,2}$.

3. (Enumerate all nodes in $\pi_{d,2}$ as $c_1, \ldots, c_k$ such that, in $\mathcal{N}$, $c_i$ is not an ancestor of $c_j$ if $i > j$. Let $Z_i$ be the set of nodes in $Z = \pi_{d,r} \cap \pi_{d,1}$ that, in $\mathcal{T}$, are ancestors to $c_1$, or $c_2$, $\ldots$, or $c_i$.) For each $i$, make $c_i$ a child of each node in $\{c_1, \ldots, c_{i-1}\} \cup Z_i$ and define

$$P(c_i|c_1, \ldots, c_{i-1}, Z_i) = \frac{\sum_{c_{i+1}, \ldots, c_k} P_{\mathcal{T}_c}(\pi_{d,r})}{\sum_{c_i, c_{i+1}, \ldots, c_k} P_{\mathcal{T}_c}(\pi_{d,r})}. \tag{6}$$

4. Return the resulting ID.

It is proved in the longer version of the paper that the reduced body is indeed an ID and it has the same optimal policies and optimal expected value as the body.

### 4.3 Expected values of value networks

The expected value of a value network is the sum of the expectations of all its utility functions. If one evaluates an ID using the scheme outlined at the beginning of this section, one will be left with a value network after finding optimal decision rules for all the decision nodes. The expected value of this network is the optimal expected value of the original ID.

Let $\mathcal{N}$ be a value network. If all value nodes are converted into random nodes via Cooper's transformation, then $\mathcal{N}$ becomes a BN. Denote the BN also by $\mathcal{N}$. The expected value of the value network is $\sum_{v \in V} P_{\mathcal{N}}(v=1)M_v$, where $V$ is the set of value nodes. For any value node $v$, let $\mathcal{N}_v$ be obtained from the BN $\mathcal{N}$ by pruning nodes outside $an(\{v\})$. Then $P_{\mathcal{N}}(v=1)=P_{\mathcal{N}_v}(v=1)$. Thus, the expected value can be obtained using the following procedure.

> Procedure expVal($\mathcal{N}$)
>
> 1. For each $v \in V$, obtain $\mathcal{N}_v$ from $\mathcal{N}$ by pruning nodes outside $an(\{v\})$. Compute $P_{\mathcal{N}_v}(v = 1)$ by calling BNinf($v, \mathcal{N}_v$).
> 2. Return $\sum_{v \in V} P_{\mathcal{N}_v}(v=1)M_v$.

Note that the following BN inference problems are solved:
$$P_{\mathcal{N}_v}(v=1) \text{ for each } v \in V.$$

### 4.4 An algorithm

The foregoing discussions lead to the following algorithm for evaluating IDs.

> Procedure evalID($\mathcal{N}$):
>
> 1. **While** there are decision nodes in $\mathcal{N}$
>    (a) Find the tail decision node $d$.
>    (b) $\mathcal{T}$=redTail($\mathcal{N}, d$).
>    (c) Call evalFun($\mathcal{T}, d$) to compute $e_{\mathcal{T}}(\pi_{d,r}, d)$ and $P_{\mathcal{T}_c}(\pi_{d,r})$.
>    (d) Find an optimal decision rule for $d$ via
>    $$\delta^*(\pi_{d,r}) = arg \ max_d e_{\mathcal{T}}(\pi_{d,r}, d).$$
>    (e) $\mathcal{B}$ = augBody($\mathcal{N}, d, e_{\mathcal{T}}(\pi_{d,r}, d)$).
>    (f) **If** some nodes in $\pi_d$ have parents in the downstream set $X_2$ of $\mathcal{N}$ w.r.t $d$,
>    $$\mathcal{B} = \text{redBody}(\mathcal{B}, P_{\mathcal{T}_c}(\pi_{d,r}), \mathcal{T}).$$

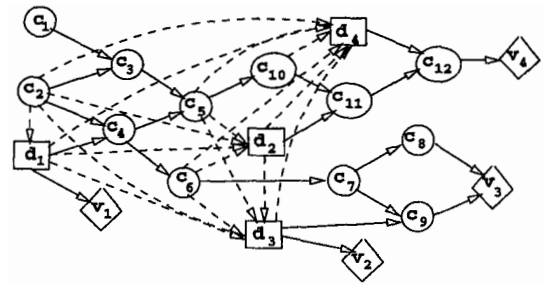

Figure 3: An ID.

> (g) $\mathcal{N}=\mathcal{B}$.
> (After the while-loop, $\mathcal{N}$ becomes a value network.)
>
> 2. Return the optimal decision rules and expVal($\mathcal{N}$).

The procedure evalID identifies a list BN inference problems and specifies how optimal decision rules can be obtained from the solutions of those problems. It leaves it to the user to choose an algorithm for solving the BN inference problems. As such, it is really an algorithm for reducing ID evaluation into BN inference problems.

## 5  An example

This section illustrates evalID by using the ID in Figure 3, which is borrowed from Jensen *et al* (1994). Arcs into decision nodes are dashed for readability.

Denote the ID by $\mathcal{N}$. Since it contains decision node, evalID enters the while-loop. In the while-loop, Step 1(a) finds that $d_4$ is the tail decision node and Step 1(b) constructs that reduced tail $\mathcal{T}$=redTail($\mathcal{N}, d_4$). To get a clear picture of $\mathcal{T}$, note that the downstream set $X_2$ is $\{d_4, c_{11}, c_{12}, v_4\}$. Since no parents of $d_2$ have parents in $X_2$, $\pi_{d_4,2}=\emptyset$. Among the parents of $d_4$, only $c_{10}$ and $d_4$ are parents to nodes in $\pi_{d_4,2} \cup X_2 \setminus \{d_4\} = X_2 \setminus \{d_4\}$, hence $c_{10}$ and $d_2$ are the all the relevant parents of $d_4$. In other words, $\pi_{d_4,r}=\{c_{10}, d_2\}$. Consequently, $\mathcal{T}$ consists of nodes $c_{10}$, $d_2$, $d_4$, $c_{11}$, $c_{12}$, and $v_4$ and is as shown in Figure 4 (2), where $v_4$ have been converted into a random node by Cooper's transformation.

Step 1(c) calculates the evaluation functional $e_{\mathcal{T}}(\pi_{d_4,r}, d_4)$. In the process, the BNs $\mathcal{T}_c$ and $\mathcal{T}_{v_4}$ are obtained from $\mathcal{T}$ by pruning nodes outside $an(\pi_{d_4,r})$ and $an(\pi_{d_4,r} \cup \{d_4, v_4\})$ respectively. Since $an(\pi_{d_4,r})=\pi_{d_4,r}=\{c_{10}, d_2\}$, $\mathcal{T}_c$ consists of two nodes $c_{10}$ and $d_2$. They are isolated from each other and both have uniform distributions. Since $an(\pi_{d_4,r} \cup \{v_4\})$ contains all nodes in the tail, $\mathcal{T}_{v_4}$ is the same as $\mathcal{T}$. The

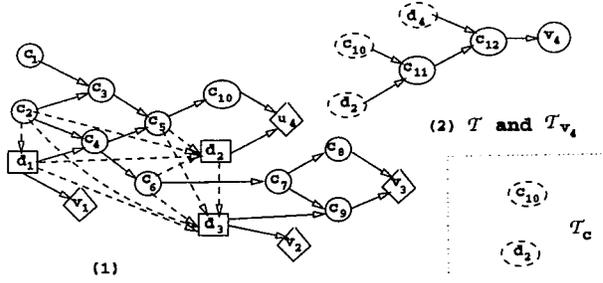

Figure 4: Reduced tail and augmented body w.r.t $d_4$.

subroutine `BNinf` is called to compute the following probabilities:

$$P_{\mathcal{T}_c}(c_{10}, d_2), P_{\mathcal{T}_{v_4}}(v_4=1, c_{10}, d_2, d_4).$$

Thereafter, the evaluation functional is obtained by

$$e_{\mathcal{T}}(c_{10}, d_2, d_4) = \frac{P_{\mathcal{T}_{v_4}}(v_4=1, c_{10}, d_2, d_4) M_{v_4}}{P_{\mathcal{T}_c}(c_{10}, d_2)/|\Omega_{d_4}|}.$$

Step 1(d) finds an optimal decision rule for $d_4$ via $\delta_4^*(c_{10}, d_2) = arg\ max_{d_4} e_{\mathcal{T}}(c_{10}, d_2, d_4)$.

Step 1(e) calls `augBody` to construct the augmented body of $\mathcal{N}$ w.r.t $d_4$, which is shown in Figure 4 (1). The utility function of the new value node $u_4$ is given by $f_{u_4}(c_{10}, d_2) = max_{d_4} e_{\mathcal{T}}(c_{10}, d_2, d_4)$. Since no nodes in $\pi_{d_4,r}$ have parents in $X_2$, Step 1(f) is skipped.

We stop here due to space limit. Interested readers are referred to a longer version of the paper for the remaining steps.

## 6    Comparisons with previous methods

Both `evalID` and the Shachter-Peot algorithm reduce ID evaluation into BN inference problems, which can be solved using arbitrary BN inference algorithms. This section shows that the probabilistic inference problems induced by `evalID` are easier to solve than those induced by the Shachter-Peot algorithm.

Among all previous algorithms, Shenoy's fusion algorithm (Shenoy 1992) and the algorithms by Ndilikilikesha (1994) and Jensen *et al* (1994) are the most efficient. Those three algorithms are basically equivalent in the sense that they all carry out essentially the same numerical computations. They are *direct evaluation algorithms* in the sense that they evaluate IDs directly without the reduction to BN inference problems. An method that reduces ID evaluation into BN inference problems would be unattractive if it is less efficient than those direct evaluation algorithms no matter what BN inference algorithm is used. We will show that this is not the case for `evalID` by comparing it with Shenoy's fusion algorithm.

Non-numerical computations in `evalID` include the identification of tail decision nodes, the construction of reduced tails and bodies, and pruning of nodes in a reduced tail that are irrelevant to particular BN inference problems . They are negligible compared to numerical computations. We will hence focus the comparisons on numerical computations.

### 6.1    Comparisons with Shachter and Peot's algorithm

The Shachter-Peot algorithm applies only when there is one value node. Let $\mathcal{N}$ be an ID with one value node. Assume that there are no barren random nodes, i.e. random nodes that have no children[3].

Let $\mathcal{N}'$ be the BN defined in Section 2. In the Shachter-Peot algorithm, the BN inference problem $P_{\mathcal{N}'}(\pi_d, d|v=1)$ needs to be solved in order to obtain an optimal decision rule for the tail decision node $d$. Let $\mathcal{T}$ be the reduced tail of $\mathcal{N}$ w.r.t $d$. In `evalID`, we need to solve two BN inference problems, namely $P_{\mathcal{T}}(\pi_{d,r})$ and $P_{\mathcal{T}}(\pi_{d,r}, d, v=1)$.

The inference problem $P_{\mathcal{N}'}(\pi_d, d|v=1)$ is more difficult to solve than $P_{\mathcal{T}}(\pi_{d,r})$ and $P_{\mathcal{T}}(\pi_{d,r}, d, v=1)$ than for two reasons. First, it involves more variables. Due to the no-forgetting constraint, all other decision nodes and their parents must be parents of $d$, i.e. in $\pi_d$. However, as we have seen in the example of the previous section, many of the parents of $d$ are irrelevant to $d$. The set $\pi_{d,r}$ usually contains much less variables than $\pi_d$. Second, $\mathcal{N}'$ usually consists of many more nodes than the reduced tail $\mathcal{T}$. Moreover, since $\mathcal{N}$ has no barren random nodes, no nodes in $\mathcal{N}'$ can be pruned when computing $P_{\mathcal{N}'}(\pi_d, d|v=1)$, while $\mathcal{T}$ can be further reduced to $\mathcal{T}_c$ when computing $P_{\mathcal{T}}(\pi_{d,r})$ and $\mathcal{T}_v$ when computing $P_{\mathcal{T}}(\pi_{d,r}, d, v=1)$.

In `evalID`, nodes in the downstream set $X_2$ are pruned after an optimal decision rule for $d$ has been obtained. In other words, those nodes do no participate in the computations for other decision nodes. However, no nodes are pruned in the Shachter-Peot algorithm. Computations for each decision node involve all nodes in $\mathcal{N}$.

### 6.2    Comparisons with Shenoy's fusion algorithm

This subsection first introduces a variation of `evalID`, called `evalID1`, that performs essentially the same

---

[3]Barren random nodes, if exist, can be pruned at a preprocessing step (Shachter 1986).

numerical computations as Shenoy's fusion algorithm and then compares evalID with evalID1.

### 6.2.1  A variation of evalID

We will refer to non-negative functions of a set of variables simply as *factors*. Conditional probabilities and utility functions are all factors. Let $\rho$ be an ordering of the nodes in $X_2 \backslash V_2 \cup \{d\}$. In stead of evalFun, evalID1 uses the following procedure to compute the evaluation functional of the reduced tail $\mathcal{T}$ and the marginal probability $P_\mathcal{T}(\pi_{d,r})$.

> Procedure evalFun1($\mathcal{T}, d$)
>
> 1. Let $\mathcal{F}$ be the list of utility functions of value nodes in $V_2$ and let $\mathcal{P}$ be the list of conditional probabilities of nodes in $X_2 \backslash V_2 \cup \{d\}$.
> 2. **While** $\rho \neq \emptyset$, remove the first node $x$ from $\rho$ and call fuse($\mathcal{P}, \mathcal{F}, x$).
> 3. Multiply all factor in $\mathcal{P}$ (the product is $P_\mathcal{T}(\pi_{d,r}, d)$.)
> 4. Sum all factors in $\mathcal{F}$ (the result is $e_\mathcal{T}(\pi_{d,r}, d)$).
> 5. Return $e_\mathcal{T}(\pi_{d,r}, d)$ and $\sum_d P_\mathcal{T}(\pi_{d,r}, d)$.

The subroutine fuse is given as follows:

> Procedure fuse($\mathcal{P}, \mathcal{F}, x$)
>
> 1. Remove from $\mathcal{P}$ all the factors $p_1, \ldots, p_k$ that involve $x$. If such factors exist, add the new factor $p = \sum_x \prod_{i=1}^{k} p_i$ to $\mathcal{P}$.
> 2. Remove from $\mathcal{F}$ all the factors $f_1, \ldots, f_l$ that involve $x$. If such factors exist, add the new factor $\sum_x [\sum_{i=1}^{l} f_i][\prod_{i=1}^{k} p_i]/p$ to $\mathcal{F}$.

Let $\mathcal{N}$ be a value network and $\rho$ be an ordering of the random nodes. Instead of expVal, evalID1 uses the following procedure to compute the expected value of $\mathcal{N}$.

> Procedure expVal1($\mathcal{N}$):
>
> 1. Let $\mathcal{F}$ be the list of utility functions of value nodes in $\mathcal{N}$ and let $\mathcal{P}$ be the list of conditional probabilities of random nodes in $\mathcal{N}$.
> 2. **While** $\rho \neq \emptyset$, remove the first node $x$ from $\rho$ and call fuse($\mathcal{P}, \mathcal{F}, x$).
> 3. Return the sum all factors in $\mathcal{F}$ (which is the expected value of $\mathcal{N}$).

Shenoy's fusion algorithm requires an ordering, say $\rho_s$, of all the random and decision nodes. If the orderings in evalFun1 and expVal1 *conforms* to $\rho_s$ in the sense that relative orders of nodes are the same, than evalID1 carries out essentially the same numerical computations as Shenoy's fusion algorithm (see Shenoy 1992).

### 6.2.2  Comparisons

We now set out to compare evalID1 and evalID. Suppose evalID employs the VE (variable elimination) algorithm [4] for probabilistic inference. Let $A$ be a subset of nodes in a BN $\mathcal{M}$ and let $\rho$ be an ordering of nodes outside $A$. VE computes the marginal probability $P_\mathcal{M}(A)$ as follows:

> Procedure BNinf-VE($\mathcal{M}, A$):
>
> 1. Let $\mathcal{P}$ be the list of conditional probabilities in $\mathcal{M}$ .
> 2. **While** $\rho \neq \emptyset$, remove the first node $x$ from $\rho$ and remove from $\mathcal{P}$ all the factors $f_1, \ldots, f_k$ that involve $x$. If such factors exist, add the new factor $\sum_x \prod_{i=1}^{k} f_i$ to the list $\mathcal{P}$.
> 3. Multiply all factors in $\mathcal{P}$ and return the result (which is $P_\mathcal{M}(A)$).

Consider computing $P_{\mathcal{T}_c}(\pi_{d,r})$ and $P_{\mathcal{T}_v}(\pi_{d,r}, d, v{=}1)$ using BNinf-VE. If the ordering in BNinf-VE conforms to the ordering in evalFun1, BNinf performs no more numerical computations than evalFun1($\mathcal{T}, d$). Therefore, the amount of numerical computations carried out by evalFun($\mathcal{T}, d$) is at most $1{+}m$ times that carried by evalFun1($\mathcal{T}, d$), where $m$ is the number of value nodes in $\mathcal{T}$. Similarly, for any value network $\mathcal{N}$, the amount of numerical computations carried out by expVal($\mathcal{N}$) is at most $m$ times that carried by expVal1($\mathcal{N}$), where $m$ is the number of value nodes in $\mathcal{N}$.

We argue that evalFun($\mathcal{T}, d$) is usually more efficient that evalFun1($\mathcal{T}, d$), especially when $\mathcal{T}$ is large. Define the *size* of a factor to be the number of variables involved in the factor. It is well understood in the BN literature that the complexities of BNinf and evalFun1 is largely determined by the sizes of the largest factors encountered; they are exponential in the largest factors sizes. The BNs $\mathcal{T}_c$ and each $\mathcal{T}_v$ are subnetworks of $\mathcal{T}$. When $\mathcal{T}$ is large, the differences between $\mathcal{T}$ and $\mathcal{T}_c$ or $\mathcal{T}_v$ are usually also large. In such a case, the maximum factor sizes encountered by BNinf are smaller than those encountered by evalFun1 and hence the amount of time BNinf-VE spends in computing $P_{\mathcal{T}_c}(\pi_{d,r})$ or $P_{\mathcal{T}_v}(\pi_{d,r}, d, v{=}1)$ is much less than that evalFun1$(\mathcal{T}, d)$ takes. Consequently, evalFun$(\mathcal{T}, d)$ takes less time than evalFun1$(\mathcal{T}, d)$.

---

[4]The idea behind VE is implicit in many papers (e.g. Shenoy 1992). It was first made explicit in Zhang and Poole (1994) and extended to exploit independence of causal influence by Zhang and Poole (1996).

A second reason for evalFun$(\mathcal{T}, d)$ being more efficient than evalFun1$(\mathcal{T}, d)$ is the fact that the former does not perform numerical divisions until the last step, while the latter might divide factors when fusing each node.

Similarly, expVal is usually more efficient than expVal1. Hence evalID is usually more efficient than evalID1 and therefore more efficient than Shenoy's fusion algorithm.

We would like to emphasize that arbitrary BN inference algorithms can be used in evalID, while this is not the case in Shenoy's fusion algorithm and all the direct evaluation algorithms for that matter. This is a big advantage (see discussions in the next section).

## 7    Conclusions

This paper is about reducing ID evaluation into BN inference problems. Such an exercise is interesting because it allows the use of arbitrary BN inference algorithms in evaluating IDs. Two reduction methods have been proposed previously (Cooper 1988 and Shachter and Peot 1994). A new method is presented in this paper. The BN inference problems induced by the new method are easier to solve than those induced by earlier methods.

When coupled with the VE algorithm, the performance of the new method is, in the worst case, within a small constant factor of that of the most efficient previous algorithms, which evaluate ID directly without the reduction into BN inference problems. We have argued that the combination of the new method and VE is usually more efficient in large IDs.

The fact that it allows arbitrary BN inference algorithms is big advantage of the new method. From a system development point of view, the method enables one to easily add ID evaluation capabilities to any BN inference packages. From the efficiency point of view, speeding up inference in BNs has been and still is an active research area. There are algorithms that exploit independence of causal influence (e.g. Zhang and Poole 1996) and that exploit special structures in the conditional probability tables. The new method facilitates ready incorporation of those algorithms, as well as future advances in BN inference, in ID evaluation. We are not aware of any approximate algorithms for IDs, while there is a rich collection of approximate and simulation algorithms for BNs. The new method also opens up the possibility of approximate algorithms for ID, which might be necessary in order to solve large decision problems.